\documentclass[10pt,a4paper]{article}
\usepackage{amsmath, amsfonts, amsthm, amssymb}
\usepackage[a4paper,margin=1in]{geometry}
\usepackage{anysize,graphicx,fancyhdr}
\usepackage{mathptmx}
\usepackage{mathtools}
\usepackage{caption}
\usepackage{authblk}
\usepackage{url}

\marginsize{2.5cm}{2.5cm}{2.5cm}{2.5cm}
\newtheorem{theorem}{Theorem}

\usepackage{sectsty}
\sectionfont{\fontsize{12}{10}\selectfont}
\subsectionfont{\fontsize{10}{8}\selectfont}

\title{\Large \bf Quantitative Risk Indices for Autonomous Vehicle Training Systems}
\date{}
\author[1]{Eduardo Candela}
\author[2]{Yuxiang Feng}
\author[3]{Panagiotis Angeloudis}
\author[4]{Yiannis Demiris}
\affil[1]{\normalsize CTS, Department of Civil and Environmental Engineering, Imperial College London\protect\\ e.candela-garza19@imperial.ac.uk}
\affil[2]{\normalsize CTS, Department of Civil and Environmental Engineering, Imperial College London\protect\\ y.feng19@imperial.ac.uk}
\affil[3]{\normalsize CTS, Department of Civil and Environmental Engineering, Imperial College London\protect\\ p.angeloudis@imperial.ac.uk}
\affil[4]{\normalsize ISN, Department of Electrical and Electronic Engineering, Imperial College London\protect\\ y.demiris@imperial.ac.uk}
\begin{document}

\maketitle
\thispagestyle{fancy}

\noindent\text{Keywords: }\textit{autonomous vehicles, risk indices, quantitative, safety, simulation, artificial intelligence.}

\bigskip

\section{Introduction}

\subsection{Problem Statement}

The development of Autonomous Vehicles (AV) presents an opportunity to save and improve lives, and also increase the quality and efficiency of transportation systems. Road safety is one of the main public health issues; traffic-related deaths are approximately 1.35 million every year according to the World Health Organization \cite{who2020}, and are the leading global cause of deaths for children and young adults. According to the NHTSA, human error is linked to 94\% of serious crashes \cite{nhtsa2017}.
  
In the past two decades, the AV field has experienced significant progress. The Defense Advanced Research Projects Agency (DARPA) in the United States sponsored a series of competitions between 2004 and 2007 that accelerated research to near-real-world conditions \cite{buehler2009darpa}. Moreover, the recent advancement in Artificial Intelligence (AI) technologies has significantly aided in the development of self-driving methods. In particular, perception with computer vision (which is essential for AV) has improved rapidly since the deep learning revolution began with AlexNet in 2012 \cite{krizhevsky2012imagenet}. However, state-of-the-art computer vision systems have still yet to achieve the required error rates for autonomous driving \cite{schwarting2018planning}.

Achieving fully capable AV will require overcoming many technical challenges. Most of the current “self-driving” commercial vehicles only present SAE Level 2 autonomy. This level means the driver must be ready to take control of the vehicle at all times. One of the main obstacles in reaching SAE Level 5 (full) autonomy is the presence of edge-cases: unsafe scenarios that happen with extremely low probabilities and make field testing infeasible. It would require driving a car during a prohibitive amount of time – in the order of hundreds of years – to ensure they are properly handled \cite{hussain2018autonomous}. Hence the necessity of simulation in training and validating the safety of AV control systems. Nevertheless, there is a gap in the literature regarding the measurement of safety for self-driving systems. Measuring safety and risk is paramount for the generation of useful simulation scenarios. 

\subsection{Research Objectives}

There is extensive research on vehicle accidents focusing on qualitative aspects and driver behavioral factors. For some examples, see \cite{hayward1972near}, \cite{klauer2006impact} and \cite{uchida2010investigation}. Regarding quantitative studies on vehicle accidents, {\AA}sljung et al. used Extreme Value Theory to estimate the frequency of collisions with near-collision data \cite{aasljung2017using}. Wang et al. trained a tree-based machine learning model with near-crash data to do driving risk assessment \cite{wang2015driving}. The limitation of these approaches is the dependence on near-crash data. Although near-miss data can substantially increase available accident data (actual crash data is very limited), the definition of a near-miss or near-crash is arbitrary – usually defined as a circumstance with a "sudden" maneuver, and labeled manually by humans. 

A promising alternative is the introduction of the Responsibility-Sensitive Safety (RSS) model by Shalev-Shwartz et al. \cite{shalev2017formal}. It consists in the definition of safe lateral and longitudinal distances that can guarantee safety under reasonable assumptions for the other vehicles' accelerations and trajectories. We present a framework that extends the RSS model for cases when reasonable assumptions or safe distances are violated, and measures the likelihood of an accident happening with risk indices that also take into account risk aversion. The use of these accident likelihood metrics can aid in the generation and selection of simulation scenarios for training and safety validation purposes. Moreover, we implemented and tested the risk indices with various multiple-vehicle settings using the simulation engine Unity.

\section{Methodology}

\subsection{Risk Index Formulation}

We introduce a single risk index $r\in[0,1]$ that represents the likelihood of a collision happening between any two vehicles moving on a two-dimensional surface. This index is the product of a longitudinal risk index $r_{lon}\in[0,1]$, and a lateral risk index $r_{lat}\in[0,1]$, which are piecewise linear functions that depend on the definitions of safe longitudinal and lateral distances (respectively) by Shalev-Shwartz et al. \cite{shalev2017formal}. The minimum safe longitudinal distance is 

\[
	d^{lon}_{min} = \left[ v_{r}\rho + \frac{1}{2}\rho^{2}a_{max,accel} + \frac{(v_{r} 
	+ \rho a_{max,accel})^{2}}{2a_{min,brake}} - \frac{v_{f}^{2}}{2a_{max,brake}} \right]_{+},
\] 
where $[x]_{+}:=\text{max}\{x,0\}$; $c_{r}$ is a car with velocity $v_{r}$ that drives behind another car $c_{f}$ (in the same direction) with velocity $v_{f}$. For any braking of $c_{f}$ of at most $a_{max,brake}$, $c_{r}$ has a response time of $\rho$ during which it accelerates by at most $a_{max,accel}$, and immediately after the response starts braking by at least $a_{min,brake}$ until necessary. Notice that the closer the two vehicles are to colliding, $c_{r}$ will have to brake harder to avoid crashing, until its brakes reach their maximum capability (which we denote $A_{brake}$) and a collision is unavoidable. When replacing $a_{min,brake}$ with $A_{brake}$ in $d^{lon}_{min}$, a tighter bound is defined for the safe longitudinal distance as 

\[
	d^{lon}_{min,brake} = \left[ v_{r}\rho + \frac{1}{2}\rho^{2}a_{r,\rho} + \frac{(v_{r} 
	+ \rho a_{max,accel})^{2}}{2A_{brake}} - \frac{v_{f}^{2}}{2a_{max,brake}} 
	\right]_{+}.
\] 
Let $d^{lon}$ be the current longitudinal distance between cars. We define the longitudinal risk index

\[
  r_{lon}=\left\{
  \begin{array}{@{}ll@{}}
    0, & \text{if}\ d^{lon} \geq d^{lon}_{min} > 0 \\
    1-\frac{d^{lon} - d^{lon}_{min,brake}}{d^{lon}_{min} - d^{lon}_{min,brake}}, & \text{if}\ d^{lon}_{min} \geq d^{lon} \geq d^{lon}_{min,brake} > 0 \\
    1, & \text{otherwise} \\
  \end{array}\right.\qquad.
\]

Similarly, for the lateral analysis the safe lateral distance is 

\[
	d^{lat}_{min} = \left[ \frac{v_{left} + v_{left,\rho}}{2}\rho 
	+ \frac{v^{2}_{left,\rho}}{2a^{lat}_{min,brake}}
	- \left(\frac{v_{right} + v_{right,\rho}}{2}\rho 
	- \frac{v^{2}_{right,\rho}}{2a^{lat}_{min,brake}} \right)
	\right]_{+},
\] 
where w.l.o.g. car $c_{left}$ is to the left of $c_{right}$, $v_{left}$ and $v_{right}$ are their respective lateral velocities, $\rho$ is the same response time as in the longitudinal case, during which the two cars will apply a maximum lateral acceleration of $a^{lat}_{max,accel}$ toward each other, and after that both cars will apply lateral braking of at least $a^{lat}_{min,brake}$ until necessary; $v_{left,\rho} = v_{left} + \rho a^{lat}_{max,accel}$ and $v_{right,\rho} = v_{right} - \rho a^{lat}_{max,accel}$. As with the safe longitudinal distance, when replacing $a^{lat}_{min,brake}$ with $A^{lat}_{brake}$ (the maximum capable lateral braking of the vehicles) in $d^{lat}_{min}$, a tighter bound is defined for the safe lateral distance as 

\[
	d^{lat}_{min,brake} = \left[ \frac{v_{left} + v_{left,\rho}}{2}\rho 
	+ \frac{v^{2}_{left,\rho}}{2A^{lat}_{brake}}
	- \left(\frac{v_{right} + v_{right,\rho}}{2}\rho 
	- \frac{v^{2}_{right,\rho}}{2A^{lat}_{brake}} \right)
	\right]_{+}.
\]

Let $d^{lat}$ be the current lateral distance between cars. We define the lateral risk index

\[
  r_{lat}=\left\{
  \begin{array}{@{}ll@{}}
    0, & \text{if}\ d^{lat} \geq d^{lat}_{min} > 0 \\
    1-\frac{d^{lat} - d^{lat}_{min,brake}}{d^{lat}_{min} - d^{lat}_{min,brake}}, & \text{if}\ d^{lat}_{min} \geq d^{lat} \geq d^{lat}_{min,brake} > 0 \\
    1, & \text{otherwise} \\
  \end{array}\right.\qquad.
\]

Multiplying the lateral and longitudinal risk indices, and adding risk propensity parameters $\beta$, $\gamma > 0$ the unified risk index is

\[
	r = \left(r_{lon}\right)^{\beta}\left(r_{lat}\right)^{\gamma},
\] 
which clearly lies in the range $[0,1]$ as well, and only takes a value $>0$ when both longitudinal and lateral risk indices are also $>0$ simultaneously. The larger the risk propensity parameters are, the smaller the risk index becomes. When the propensity parameters are $<1$, the risk propensity becomes a risk aversion, since the longitudinal and lateral risk indices are increased.

\begin{theorem}
\label{t1}
When $r=0$, a collision is impossible to take place in the future given the current states of the vehicles (i.e. positions, velocities and accelerations). If $r=1$, avoiding a potential collision cannot be guaranteed under the assumptions of safe longitudinal and lateral distances. This is independent of the risk propensity parameters $\beta$ and $\gamma$.
\end{theorem}

\subsection{Expected Results}

The risk index $r$ formulated in the previous subsection is currently being implemented and tested with multiple car-following settings and scenarios using the simulation engine Unity. In order to test the validity and usefulness of the proposed index, two sets of experiments will be conducted:

\begin{table}[h!]
\begin{tabular}{|c|l|}
\hline
\textbf{Experiment}      & Random changes in driving behavior                                                  \\ \hline
\textbf{Description}     & Reasonable accelerations and trajectories of traffic vehicles are randomly violated. \\ \hline
\textbf{Reason}     & Capture human error. \\ \hline
\textbf{Expected result} & Positive relationship between risk index increase and simulated collisions.         \\ \hline
\end{tabular}
\end{table}

\begin{table}[h!]
\begin{tabular}{|c|l|}
\hline
\textbf{Experiment}      & Random changes in vehicle state estimation                                           \\ \hline
\textbf{Description}     & State variables are randomly perturbed for ego vehicle as well as traffic vehicles. \\ \hline
\textbf{Reason}     & Capture sensor and perception errors. \\ \hline
\textbf{Expected result} & Positive relationship between risk index increase and simulated collisions.         \\ \hline
\end{tabular}
\end{table}

\section{Conclusions}

Autonomous Vehicles (AV) have experienced an accelerated development in the past two decades. Some of the main factors of this recent development are the resurgence and explosive growth of Artificial Intelligence since 2012, military investment (DARPA challenges in 2004-2007), and increased interest and involvement of automotive and software industries. Nevertheless, the state of the art on self-driving vehicles still presents limitations that must be solved in order to reach SAE Level 5, i.e. full autonomy. 

Validation of AV control systems by field testing is intractable, because it could require up to hundreds of years of driving due to the heavy-tail distribution of safety edge-case appearances over time. Therefore, safety validation by design and also by simulation are essential for the development and successful launch of AV. There is a gap in the literature regarding the definition of quantitative safety metrics that allow a more efficient generation of simulation scenarios for both training and validation of self-driving vehicles.

In this paper, we introduce quantitative risk indices that capture the increased likelihood of an accident happening between two vehicles given their physical states – this can be extended to cases with multiple vehicles. The presented indices were also implemented and tested with multiple car-following settings using the simulation engine Unity.

The use of this new indices can significantly aid in the creation of automated algorithms for generation of more efficient simulation scenarios for the training of AV control systems, and also their safety validation. Moreover, we remove the dependency on arbitrary near-crash data used in other studies, which can be biased. We identify and suggest four possibilities for future research based on the presented framework: 1) the development of a cost function for scenario search, 2) the use of AI generative techniques such as GANs for scenario generation, 3) further testing and calibration of the indices using real data, 4) the definition of additional risk indices that capture more specific accident settings.

\bibliographystyle{unsrt} 
\bibliography{biblio}

\end{document}